# FunduSegmenter: Leveraging the RETFound Foundation Model for Joint Optic Disc and Optic Cup Segmentation in Retinal Fundus Images

*Zhenyi Zhao, Muthu Rama Krishnan Mookiah, Emanuele Trucco*

VAMPIRE project,
Computing, School of Science and Engineering
University of Dundee
Dundee, UK

Corresponding author: Zhenyi Zhao, 2578745@dundee.ac.uk

## Abstract
Purpose: This study introduces the first adaptation of RETFound for joint optic disc (OD) and optic cup (OC) segmentation. RETFound is a well-known foundation model developed for fundus camera and optical coherence tomography images, which has shown promising performance in disease diagnosis.

Methods: We propose FunduSegmenter, a model integrating a series of novel modules with RETFound, including a Pre-adapter, a Decoder, a Post-adapter, skip connections with Convolutional Block Attention Module and a Vision Transformer block adapter. The model is evaluated on a proprietary dataset, GoDARTS, and four public datasets, IDRiD, Drishti-GS, RIM-ONE-r3, and REFUGE, through internal verification, external verification and domain generalization experiments.

Results: An average Dice similarity coefficient of 90.51% was achieved in internal verification, which outperformed all baselines, some substantially (nnU-Net: 82.91%; DUNet: 89.17%; TransUNet: 87.91%). In all external verification experiments, the average results were about 3% higher than those of the best baseline, and our model was also competitive in domain generalization.

Conclusions: This study explored the potential of the latent general representations learned by RETFound for OD and OC segmentation in fundus camera images. Our FunduSegmenter generally outperformed state-of-the-art baseline methods. The proposed modules are general and can be extended to fine-tuning other foundation models.

Translational Relevance: The model shows strong stability and generalization on both in-distribution and out-of-distribution data, providing stable OD and OC segmentation. This is an essential step for many automated tasks, from setting the accurate retinal coordinate to biomarker discovery.

The code and trained weights are available at: https://github.com/JusticeZzy/FunduSegmenter.

## 1. Introduction

Changes in the appearance of anatomical structures of the retina observed in color fundus photography, mainly optic disc (OD), optic cup (OC), fovea and above all retinal vessels have been identified as promising sources of biomarkers for systemic diseases [1-14]. Segmenting such structures accurately is essential for retinal biomarker research.

RETFound [15] is recognized as the first foundation model (FM) built from a large collection of retinal fundus camera and optical coherence tomography (OCT) images. Its effectiveness has been demonstrated on stratification tasks based on diagnosing retinal and systemic diseases from retinal images [16-18].

We propose FunduSegmenter, to our best knowledge, the first adaptation of RETFound for a substantially different task, in our case OD and OC segmentation in fundus camera images. We compare results against several state-of-the-art baselines and segmentation-specific architectures. We find that our FunduSegmenter outperforms baseline networks on both in-distribution and out-of-distribution data, showing strong stability and generalization capability. We comment on these results in the discussion section.

Supervised Convolutional Neural Network (CNN) models have shown strong performance in retinal image segmentation [1-9]. However, at least three limitations remain. First, unlike natural images, annotating medical images, a notoriously labor-intensive and time-consuming task, requires expert clinicians. For this reason, many medical image repositories remain unlabeled and unexploited. Second, the performance of supervised CNN models decreases rapidly when tested on out-of-distribution datasets, and domain generalization methods proposed recently [19-21] need large amounts of data. Third, almost all supervised CNN models implement data augmentation with either basic transformations or by generating completely new synthetic images (i.e., not modifications of existing ones). The former does not add much new information; the latter does, but AI systems trained with image collections containing large amounts of synthetic images (phantoms) are trusted less by clinicians than those trained only or mostly with real images [22, 23].

FMs have been introduced recently [15, 24-28] and address the problems above. They are normally trained on very large volumes of unlabeled data using self-supervised learning (SSL) to generate a latent representation of a *whole* image domain. The representation is then used to perform specific downstream tasks after training a head with a limited amount of task-specific labelled data [24]. An at-a-glance comparison of this paradigm with traditional, task-specific supervised learning models is shown in Figure 1.

RETFound is based on masked autoencoders (MAE) [25], a well-known SSL technology originally demonstrated on natural images with the ImageNet-1K dataset [29]. The MAE encoder is built in RETFound as the pre-trained encoder. Two independent RETFound models were trained on 904,170 fundus camera images and 736,442 OCT images [15]. As proven by RETFound and other independent FM work [25-28], FMs address the limitations of supervised CNN models: first, unlabeled data are used effectively; second, generalization is improved significantly; third, RETFound, trained fully on real retinal images with only basic geometric augmentations, is more acceptable to clinicians than systems relying on augmentation with synthetic phantoms.

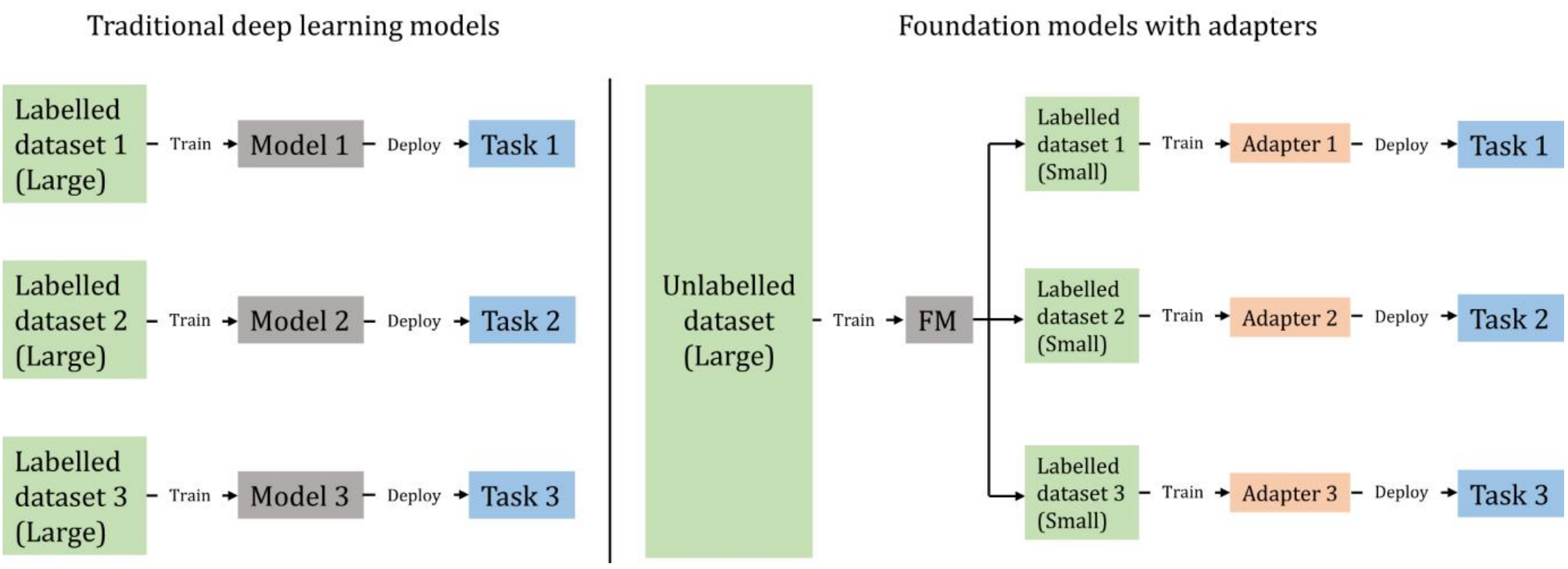

*Figure 1. Differences between traditional deep learning models (left) and FMs with adapters (right).*

To our best knowledge, RETFound has not yet been adapted for segmenting anatomical structures in fundus camera images. We explore this adaptation in FunduSegmenter focusing on segmenting the OD and OC in fundus camera images. We adopt the encoder of RETFound as the pre-trained encoder and the decoder of Segmenter [30] as the decoder; we also introduce a novel adaptation architecture composed of a series of modules, including a Pre-adapter, a Post-adapter, skip connections with Convolutional Block Attention Module (CBAM) [31] and a Vision Transformer (ViT) block adapter. The proposed Pre-adapter accommodates different sizes for input images; the Post-adapter provides upsampling stability during training; skip connections exploit multi-scale information existing in the RETFound representation; and the ViT block adapter achieves a better segmentation adaptation for the RETFound encoder. The total amount of tunable parameters is 35.57M, which is competitively small. We trained the model using annotated images from a local dataset (GoDARTS [32]) and four public datasets (IDRiD [11], Drishti-GS [12], RIM-ONE-r3 [13], REFUGE [14]). The evaluation experiments include internal verification, external verification and domain generalization, offering a comprehensive evaluation of in-distribution and out-of-distribution generalization. We compared performance with six state-of-the-art baseline systems: VAMPIRE's DUNet [1] (the segmentation system adopted in the VAMPIRE software system [10]), nnU-Net [33], TransUNet [34], the benchmark domain generalization model for joint OD and OC segmentation, DoFE [19], and two further state-of-the-art domain generalization models, RAM-DSIR [20] and TVConv [21].

In summary, the contributions of this paper are as follows.

1. To our best knowledge, the first use of RETFound for retinal image segmentation.
2. Achieving better performance *overall* than all baselines, and nearly all (remaining competitive) in internal verification, external verification and domain generalization experiments.
3. The proposed modules require no prior knowledge and can be extended to other FMs with ViT as the encoder trained through SSL.

## 2. Related Work

Here we introduce further related work in addition to what discussed in Section 1 on CNNs for OD and OC segmentation.

### 2.1. Fine-tuning on SSL FMs

SSL FMs like MAE [25], BEiT [26], SimMIM [27] and UM-MAE [28] have shown strong generalization capability on classification tasks for both natural images and medical images. However, research on their adaptation to segmentation tasks is still limited [35-37]. Although the original papers reported segmentation adaptation experiments, the segmentation decoders remained the same (use UPerNet [38] or SETR [39]) and the work focused on improving the encoder and SSL strategies. In the medical image domain, ViT-UperNet [35] applied UPerNet to a MAE pre-trained on cardiac MRI images for a segmentation task. A modified SimMIM pre-trained on wrist ultrasound images [36] was incorporated into TransUNet and U-Net. Additionally, the MAE was extended to train on 3-dimensional images (brain MRI) [37], and incorporated into a nnU-Net. These models proved that SSL FMs can be adapted to medical segmentation tasks, but the focus remained on upgrading the encoder and SSL strategies. Research is still lacking on achieving better adaptation for segmentation with SSL FMs and leveraging strong general representations from FMs for out-of-distribution generalization after fine-tuning. Importantly for our work and to our best knowledge, there is still no work leveraging RETFound for retinal image segmentation.

### 2.2. Domain Generalization in OD and OC Segmentation

Domain generalization aims to learn representations that can be used effectively with a variety of unseen out-of-distribution data, for instance acquired by different instruments or from different clinical centers. DoFE [19] established a detailed experimental framework which serves as a domain generalization benchmark for joint OD and OC segmentation [20, 21]. To improve domain generalization, the method leverages a knowledge pool of domain sources and dynamically augments image features through domain code prediction and attention-guided feature embedding. Another state-of-the-art domain generalization model, RAM-DSIR [20], introduced a Random Amplitude Mixup (RAM) module for frequency-based augmentation, a Domain-Specific Image Restoration (DSIR) module for self-supervised regularization, and a semantic consistency loss to enhance model robustness against domain shifts. Furthermore, TVConv [21] is a Translation Variant Convolution (TVConv) operator which enhances domain generalization on images with similar layouts by generating spatially adaptive convolutional kernels through learnable affinity maps and an over-parameterized weight-generating block. We use these recent methods, which apply specific enhancements for domain generalization, as baselines for our experiments.

## 3. Methods

### 3.1. Proposed Model

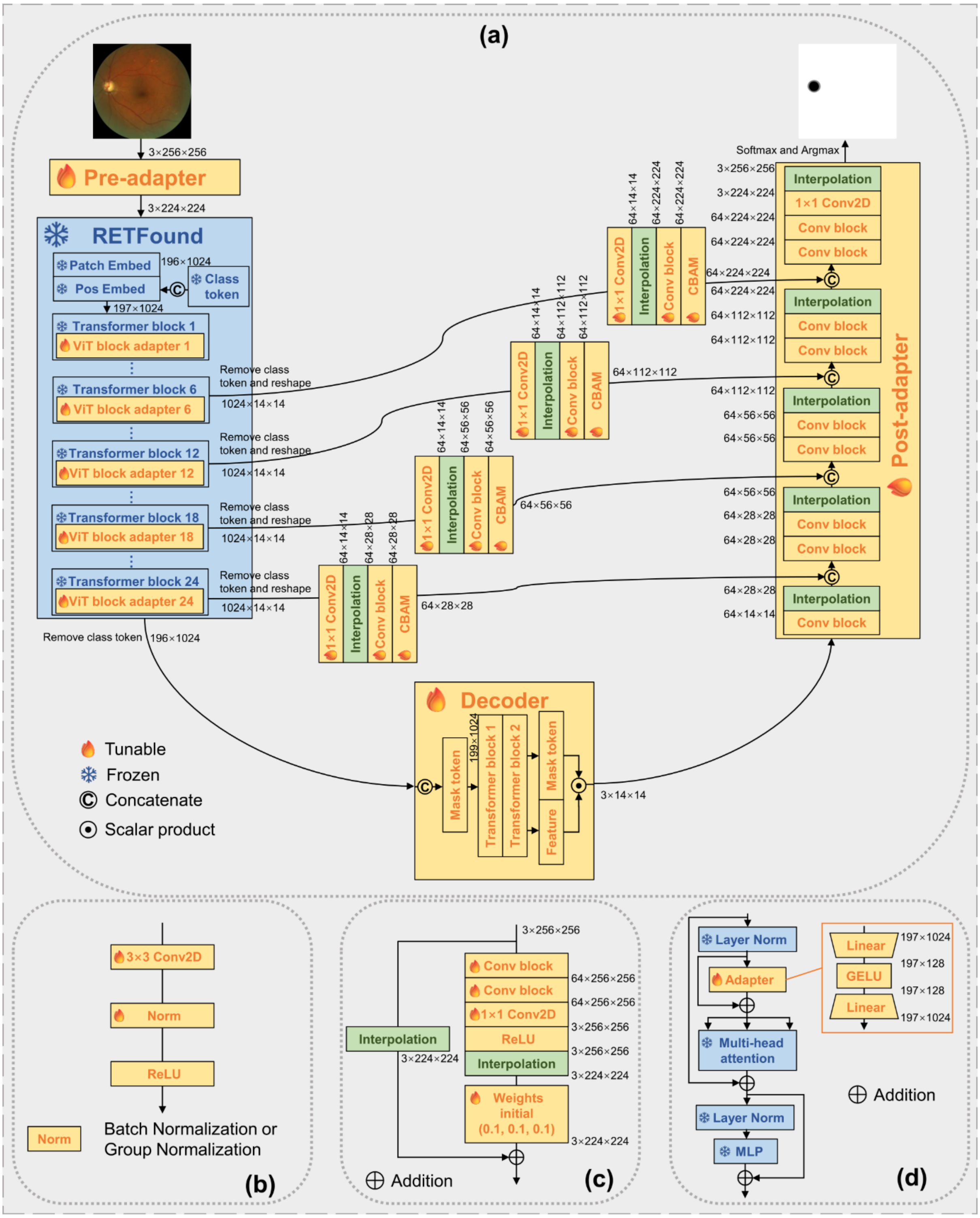

*Figure 2. Model architecture of FunduSegmenter. (a) Overall architecture. (b) The basic CNN block used in the model. (c) The architecture of the Pre-adapter. (d) The position of the ViT block adapter and its architecture.*

We adapted the decoder from Segmenter [30] for use with the fundus version of RETFound [15]. We chose the Segmenter decoder as it has been shown to outperform UPerNet [38] and SETR [39] which are widely used for segmentation adaptation of FMs with natural images [25, 26, 28]. Additionally, we propose a Pre-adapter, a Post-adapter and a ViT block adapter, and apply skip connections with CBAM [31]. The overall architecture proposed is presented in Figure 2.

#### 3.1.1. RETFound

RETFound [15] uses the large Vision Transformer, ViT-large [40], with 24 transformer blocks as the encoder component. The embedding vector size is 1024 and the number of multi-head attention heads is 16. The unlabeled input fundus images are split into 16 × 16 patches, 75% of which are randomly masked. Only unmasked patches are projected onto feature vectors. A class token is added for the image representation to enable class predictions in the downstream classification task. Position embeddings are added to all features. ViT-large takes the feature vectors as input and reconstructs the original images through a decoder. After completing this SSL process, RETFound has learned rich hidden information from the whole unlabeled set of input images. The decoder is then removed and a Multi-Layer Perceptron (MLP) layer adapted to the encoder for classification tasks.

#### 3.1.2. Our Method

*A. Segmenter Decoder*

In our work, we froze the weights of RETFound and removed both the MLP layer and the class token, as we aimed to use RETFound to extract high-level features. Then we adapted the decoder of Segmenter [30] to work with RETFound's encoder. The decoder contains a Mask Transformer block, including two transformer blocks with an embedding vector size of 1024 and 16 multi-head attention heads. We added new class tokens for masks, as many as the mask classes. We initialized the weights using values from a truncated normal distribution. The Mask Transformer takes the high-level features from RETFound and generates new feature vectors. Mask maps are obtained by the scalar product between class tokens and feature vectors. However, we removed the subsequent upsampling layer and replaced it with the proposed Post-adapter (details in Section *C. Post-adapter*).

*B. Pre-adapter*

RETFound input images are 224 × 224 pixels in size and it uses position interpolation for fine-tuning on images of different sizes. Instead, we propose a lightweight Pre-adapter (Figure 2 (c)) with 40K tunable parameters to accommodate different input sizes. Our ablation study shows that our Pre-adapter outperforms the position interpolation method (Section 4.4). We used a basic CNN block (Figure 2 (b)) containing a 2D convolutional layer with a 3 × 3 kernel, followed by a normalization layer and a ReLU activation function. The default normalization layer is batch normalization, but group normalization is provided as an alternative hyperparameter option for training with small batch size. This basic CNN block is used in both the Pre-adapter and the Post-adapter. We designed the Pre-adapter as an inverted residual architecture inspired by MobileNetV2 [41]. Two CNN blocks are stacked, followed by a 2D convolutional layer and a ReLU activation function. The number of input and output channels is 3, and the number of channels in all the middle layers is 64. Input images are processed by these CNN layers and then bicubically downsampled to 224 × 224. Additionally, a copy of the input images is directly bicubically downsampled to 224 × 224 and added to the main path through a weighted residual connection. The feature maps from the main pathway are multiplied by three independent tunable scalar weights, one for each channel, initialized at 0.1. This design allows the extraction of features from images of different sizes while minimizing the introduction of noise.

*C. Post-adapter*

In the original Segmenter decoder, the final segmentation maps are obtained by direct bilinearly upsampling of the mask maps (obtained by scalar product) to the input image size. This approach works well in natural image segmentation as the targets usually have generally clear boundaries. However, in joint OD and OC segmentation, the boundaries are less clear, especially between OD and OC, possibly causing serious segmentation errors. To address this, we propose a Post-adapter (Figure 2 (a)) to progressively upsample the mask maps to the input size. Our ablation study shows that our Post-adapter can considerably improve both convergence speed and segmentation accuracy (Section 4.4). Specifically, we first stacked four modules, each combining a basic CNN block and a 2-times-upsampling bicubic interpolation layer to upsample the mask maps from the size of 14 × 14 to 224 × 224. The first basic CNN block has 3 input channels, all others 64. Then we used a basic CNN block, an output layer (a 2D convolutional layer with a 1 × 1 kernel) and a final bicubic interpolation layer to output the mask maps at the input size. The last CNN block has 64 input and output channels; the output layer has 64 input channels, 2 output channels for OD segmentation and 3 for joint OD and OC segmentation. The placement of all the blocks (CNNs, interpolation and output layers) was determined experimentally. A softmax function produces the probability maps, and an argmax operation obtains the final binary masks.

*D. Skip Connections*

We extracted feature maps from the middle layers (6, 12, 18, and 24) of RETFound and passed them to the Post-adapter through skip connections (Figure 2 (a)). Specifically, we first copied the feature maps output from these layers and removed the class tokens. Each feature map was reshaped from its sequential form to 14 × 14 with 1024 channels, then projected to 64 channels by a 2D convolutional layer with a 1 × 1 kernel. Each projected feature map was then processed by an independent module containing a bicubic upsampling interpolation layer and a basic CNN block with 64 input and output channels. The projected feature maps from the layers mentioned are upsampled to 224 × 224, 112 × 112, 56 × 56, and 28 × 28, respectively. Finally, the upsampled feature maps are processed by four independent CBAM blocks [31] and then concatenated with feature maps of the same size in the Post-adapter. Each concatenation operation is followed by a basic CNN block with 128 input channels and 64 output channels. These skip connections elicit multi-scale information, improving in particular segmentation accuracy for the OC, which is quite smaller than the OD.

E. *ViT Block Adapter*

All the modules process the *fixed* feature maps extracted by the frozen RETFound from each image during training, limiting the performance upper bound of the model. To address this we implemented two methods, namely *partial freezing* and *ViT block adapter* (Figure 2 (d)). Our ablation study shows that partial freezing does not outperform full freezing in our architecture (Section 4.4), but the ViT block adapter improves the performance considerably. Specifically, we used a simple adapter placed between the layer normalization (LN) layer and the Multi-Head-Attention (MHA) block of each transformer block. A residual connection was applied to allow the model to skip the adapter. The adapter was designed as a bottleneck architecture containing a linear layer, followed by a GELU activation function and the other linear layer. It has 1024 input and output channels and 128 middle channels. The residual connection and bottleneck design enable the frozen RETFound encoder to both adapt to the segmentation task and retain the strong general feature representations by tuning the adapter parameters.

3.1.3. Loss Functions

We used a loss function combining Dice loss and Cross Entropy Loss (CELoss). This loss combination effectively balances pixel-level classification with region-level overlap, which is widely adopted in medical image segmentation and is implemented in the MONAI framework [42]. The definition of Dice loss is:

$$L_{Dice} = 1 - \frac{2TP}{2TP + FP + FN}, \tag{1}$$

where $TP$ is true positives, $FP$ false positives, and $FN$ false negatives, all at the pixel level. The definition of CELoss is:

$$L_{CE} = -\frac{1}{n}\sum_{i=1}^{n} y_i \log \hat{y}_i\,, \tag{2}$$

where $n$ is the number of samples, $y$ is the ground truth segmentation label, and $\hat{y}$ is the predicted label for image $i$. Thus, the combined loss function is:

$$L_{Total} = L_{Dice} + L_{CE}. \tag{3}$$

### 3.2. Experiment Design

#### 3.2.1. Experiment Plan

In order to evaluate the potential of using RETFound for joint OD and OC segmentation in fundus camera images, especially its generalization performance, we designed three experiments: *internal verification*, *external verification* and *domain generalization*.

*Internal verification*: we divided each dataset used (independently) in training / validation and testing subsets, and used the subsets as done customarily (all processing uses a single dataset).

*External verification:* we trained the model on the training and validation set of each dataset, one at a time, then tested performance on the testing sets of all the other datasets.

*Domain generalization*: following the experimental design of one of our baselines, DoFE [19], in its original paper, we trained the model using all images from three of the four training sets simultaneously; we then tested performance on the testing set of the fourth dataset, unused for training, and repeated the experiments for each dataset as testing dataset. This experiment can be considered a simplified external verification.

As described in Section 1, nnU-Net [33], DUNet [1] and TransUNet [34] were used as baselines in internal and external experiments. DoFE [19], RAM-DSIR [20] and TVConv [21] were used as baselines in domain generalization experiments. In addition, we used three data augmentation strategies, namely *None*, *Spatial* and *Designed* augmentation.

In *None,* we did not apply any data augmentation.

In *Spatial,* we applied three basic spatial-level augmentations, namely random rotation, vertical flipping and horizontal flipping.

In *Designed,* we applied five additional pixel-level augmentations as well as *Spatial*, so the total augmentations included random rotation, vertical flipping, horizontal flipping, Gaussian blur, Gaussian noise, brightness adjustment, contrast change, and gamma correction. This choice was inspired by deep stacked transformations [43], a stack of augmentations designed for medical images.

Our aim here is to evaluate the performance of FunduSegmenter under different levels of data availability from multiple datasets. The combination of the smallest dataset with the *None* augmentation represents the scarcest level, while the largest dataset with the *Designed* augmentation represents the richest level.

3.2.2. Datasets

| *Task* | *Datasets* | *Number of images* | | *Number of annotators* | *Ways to solve disagreement* | *Annotators* | *Automation* |
|---|---|---|---|---|---|---|---|
| | | Training | Testing | | | | |
| *Internal and external verification* | GoDARTS [32] | 201 | 25 | 1 | - | Retinal specialists | Semi-automatic |
| | IDRiD [11] | 54 | 27 | 2 | Discussion | Retinal specialists | Semi-automatic |
| | Drishti-GS [12] | 50 | 51 | 4 | Agreement from 3 experts | Glaucoma experts | Manual |
| | RIM-ONE-r3 [13] | 99 | 60 | 5 | Average | 4 Ophthalmologists and 1 optometrist | Manual |
| | REFUGE [14] | 800 | 400 | 7 | Majority voting | Glaucoma specialists | Manual |
| *Domain generalization* | Domain 1 (Drishti-GS) | 50 | 51 | - | - | - | - |
| | Domain 2 (RIM-ONE-r3) | 99 | 60 | - | - | - | - |
| | Domain 3 (REFUGE (train)) | 320 | 80 | - | - | - | - |
| | Domain 4 (REFUGE (validation)) | 320 | 80 | - | - | - | - |

*Table 1. OD/OC segmentation datasets, annotation characteristics and splits, defining the domains used in the main text.*

Table 1 lists the datasets and their splits in our experiments. GoDARTS and IDRiD only contain OD ground truth; the others both OD and OC ground truth. We randomly partitioned all the training sets into training and validation sets in a 4:1 ratio, except REFUGE, which has an independent validation set. The domain dataset setup followed that of DoFE, with Drishti-GS as Domain 1, RIM-ONE-r3 as Domain 2, REFUGE training set as Domain 3, and REFUGE validation set as Domain 4. Note that the way to solve the disagreement for Domain 1 (Drishti-GS) did not follow DoFE. The latter used the agreement of all experts as the ground truth [19], which produced extremely conservative masks in some cases (example in Supplementary Figure S1). Instead, we followed the original paper of Drishti-GS to use the agreement of three experts as the ground truth [12]. Following the analysis of the original paper of DoFE [19], we visualized with t-SNE [44] the relative positions in a 2D subspace of the distributions of image features extracted from all the datasets by VGG16 [45] pretrained on ImageNet [29]. The visualization results are shown in Figure 3; different datasets are color coded. The training and testing set within each dataset have similar distributions, except for GoDARTS and REFUGE. The relative positions of the distributions of the various datasets are quite different, and similar only for some images from GoDARTS and IDRiD, and a few of training samples from Drishti-GS (Figure 3, left). However, the distributions of the designed domain datasets are clearly separated from each other (Figure 3, right).

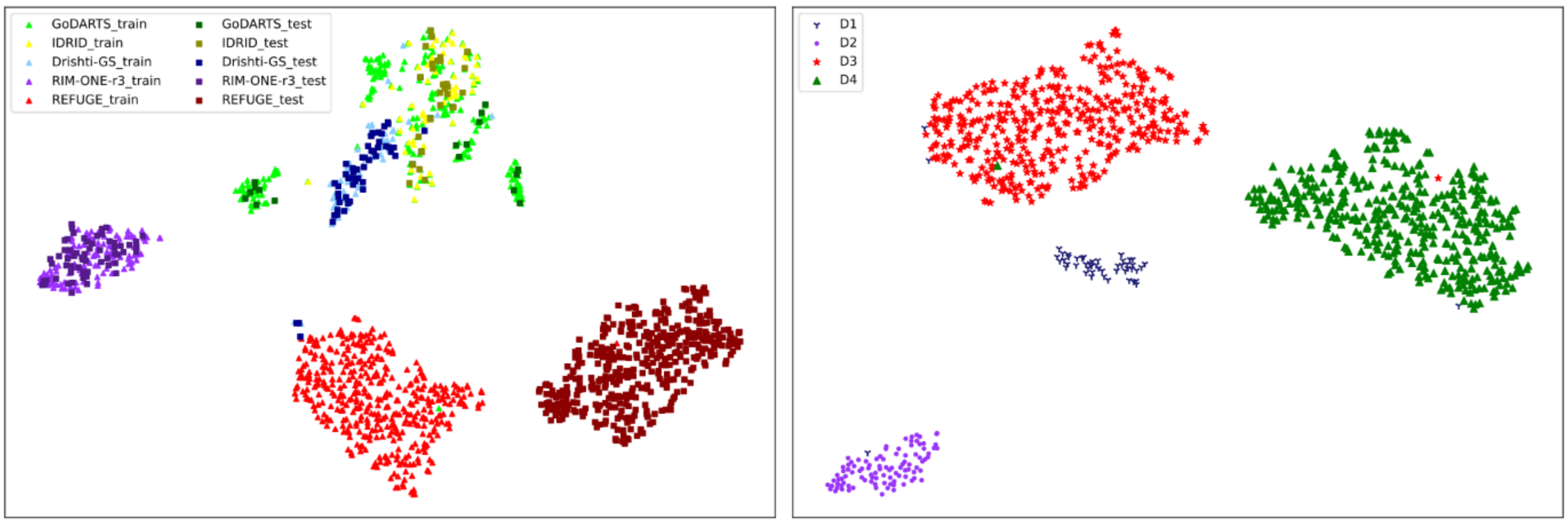

*Figure 3. Left: t-SNE visualization of image feature clusters in a 2D subspace, from training sets and testing sets of all datasets used. Right: same but for the four domain datasets defined in Table 1. D1, D2, D3 and D4 stand for Domain 1, …, Domain 4.*

These datasets also contain rich clinical information. *IDRiD:* All images are from Diabetic Retinopathy (DR) patients with various incidence of microaneurysm, soft exudate, hard exudates and hemorrhages. *Drishti-GS:* The training set contains 32 images from glaucoma patients and 18 from healthy controls. The testing set has 38 images from glaucoma patients and 13 from healthy controls. *RIM-ONE-r3:* This dataset contains 74 images from glaucoma or suspected glaucoma patients and 85 from healthy controls. *REFUGE:* In each subset, 40 images are from glaucoma patients and 360 images from healthy controls. The GoDARTS dataset, which was randomly selected from a large cohort, was used exclusively for OD segmentation in this experiment; consequently, clinical information was not retained.

#### 3.2.3. Evaluation Metrics

We adopted four metrics to evaluate the OD/OC segmentation performance, namely Dice similarity coefficient (DSC), 95% confidence interval for DSC (DSC 95% CI), 95% of the Hausdorff Distance (HD95), and Average Surface Distance (ASD). For internal and external verification, we report DSC and DSC 95% CI results; for domain generalization, DSC, HD95 and ASD results.

The definition of DSC is:

$$DSC = \frac{2TP}{2TP + FP + FN}, \tag{4}$$

where $TP$ is true positives, $FP$ false positives, and $FN$ false negatives, all at the pixel level.

DSC 95% CI was calculated through a non-parametric bootstrap approach, as some testing sets are small (e.g., GoDARTS has only 25 testing samples). We performed 10,000 resamples with replacement and calculated the percentile-based confidence interval from the resulting bootstrap distribution.

The definition of HD95 is:

$$95\%\ HD = \max\left\{h\left(\hat{Y}, Y\right), h\left(Y, \hat{Y}\right)\right\}, \tag{5}$$

where $\hat{Y}$ and $Y$ are the sets of points (pixels) of predictions and ground truth respectively, and $h\left(\hat{Y}, Y\right)$ and $h\left(Y, \hat{Y}\right)$ the Hausdorff distances from $\hat{Y}$ to $Y$ and $Y$ to $\hat{Y}$, respectively, which are defined as:

$$h\left(\hat{Y}, Y\right) = 95\%\ \max_{i\in[1,n]}\left\{\min_{j\in[1,n]}\{d(\hat{y}_i, y_j)\}\right\}, \tag{6}$$

$$h\left(Y, \hat{Y}\right) = 95\%\ \max_{j\in[1,n]}\left\{\min_{i\in[1,n]}\{d(y_j, \hat{y}_i)\}\right\}, \tag{7}$$

where $n$ is the number of points, $\hat{y}$ and $y$ are points belonging to $\hat{Y}$ and $Y$, respectively, and $d$ is the Euclidean distance (but can be other metrics).

The definition of ASD is:

$$ASD = \frac{1}{n}\sum_{i=1}^{n} \min_{j\in[1,n]}\{d(\hat{y}_i, y_j)\}, \tag{8}$$

where $n$, $\hat{y}$, $y$ and $d$ are defined as for the HD95 metric.

3.2.4. Dataset Pre-processing and Post-processing

We applied OD center cropping as pre-processing. All the images were cropped to an 800 × 800 area centered on OD, except the GoDARTS training set and IDRiD. The reason was that the size of images in the GoDARTS training set is too small (680 × 1024), and in IDRiD too large (2848 × 4288). Hence we cropped a 400 × 400 area for GoDARTS images and a 1200 × 1200 area for IDRiD ones. We pre-trained DUNet [1] on Drishti-GS, RIM-ONE-r3, the REFUGE training set and the REFUGE testing set to perform OD center cropping. For the segmentation results, we applied an area check to filter out extremely small or large segmentations and a circularity check to filter out segmentations with unreasonable shapes, ensuring the correct OD area was selected. Then we applied the minimum enclosing circle method to define a circular OD area and calculated the OD center. Note that this pre-trained DUNet weight was used only for OD center cropping and not in the main experiments. All cropped images were resized to 256 × 256 using inter-area interpolation (resampling based on pixel area relation). Three data augmentation strategies were applied (Section 3.2.1) depending on the specific task, followed by Z-score normalization.

We did not apply specific post-processing to improve segmentation results; we only bicubically upsampled segmentation results from 256 × 256 to 800 × 800 (1200 × 1200 for IDRiD) and padded them to the original image size based on the OD center locations. In domain generalization experiments, we only applied the upsampling operation, following the original DoFE paper [19].

All internal verification, external verification and domain generalization experiments of our model and baselines followed the data pre-processing and post-processing approaches described above, except those of nnU-Net. Since nnU-Net is recognized for its self-contained design integrating data pre-processing and post-processing [33], we followed its original design.

#### 3.2.5. Implementation Details

The learning rate was 0.001 and the batch size was 32. AdamW optimization was used to optimize the weights; the weight decay coefficient was 0.001. We set a large maximum number of epochs (20,000) for training. The weights corresponding to the best performance (highest DSC) on the validation set were saved as the final ones. We reported the epoch at which the best weights were obtained for each training (much smaller than 20,000). Note that only experiments of DUNet, TransUNet and our FunduSegmenter followed the implementations described above; experiments of nnU-Net and domain generalization models (DoFE, RAM-DSIR, and TVConv) followed the original implementations. The random seed was 112316 for all experiments. All experiments were run on a commercial-grade desktop with a 32GB memory size NVIDIA GeForce RTX 5090 GPU and a 4.3GHz AMD Ryzen 9 9950X CPU under Ubuntu 24.04.

### 3.3. Ablation Studies

We designed three groups of ablation experiments to comprehensively evaluate the effectiveness of each module and the performance of the final model architecture. The *Designed* augmentation strategy was used in all ablation experiments as the most comprehensive of the three mentioned in Section 3.2.1.

In brief, Group 1 investigates the effectiveness of each proposed module and aims to improve performance. Group 2 aims to improve the performance and convergence speed on small datasets. Group 3 aims to assess model generalization and stability when applying different data pre-processing approaches.

*Group 1*. First, we started with the baseline, RETFound plus Segmenter decoder, and tried decoders with 2, 4, and 6 layers. REFUGE was used as the dataset. We also tried unfreezing the last two layers of RETFound. Second, we added the Post-adapter to the best configuration from the previous step. Third, we added the Pre-adapter to the best configuration from the previous step. We tried input sizes of 224 × 224, 464 × 464, and 320 × 320. The 224 × 224 size was used to compare models with and without Pre-adapter, matching RETFound's fixed input size. The 464 × 464 size was used to compare Pre-adapter against the position interpolation method, as this was the largest size feasible for position interpolation under our experiment pipeline. The 320 × 320 size was used to compare Pre-adapter against position interpolation with fine-tuned position embeddings, which was the largest possible size. We also tried unfreezing the last two layers of RETFound in these comparisons. Additionally, we presented the results of 256 × 256, the size chosen finally. Fourth, we added the skip connections with CBAM to the best configuration from the previous step. Note that we switched to using Drishti-GS instead of REFUGE from this step, as the performance began to remain from this step on a large dataset like REFUGE (400 training images), but continued to improve significantly on a small dataset like Drishti-GS (50 training images). Fifth, we added the ViT block adapter to the best configuration from the previous step. We tried inserting it at two key positions: one between the LN and the Multi-layer Perceptron (MLP), and the other between the LN and MHA, the latter being our final model.

*Group 2.* When training the baseline (RETFound plus Segmenter decoder), delayed abrupt convergence always happened, especially on small datasets with OD only masks (e.g., IDRiD). This means the loss remained at a high or suboptimal level for tens of thousands of epochs and abruptly converged to high-accuracy generalization levels before showing any generalization evidence. Our proposed modules solved this problem and also increased model convergence speed dramatically. We conducted a mini group of ablation experiments on IDRiD with the best architecture from each step in Group 1 to present the effectiveness.

*Group 3.* The training set image size of GoDARTS is quite small (680 × 1024) and differs substantially from the testing set size (1958 × 2588). Although we cropped only a 400 × 400 area for the training set, the OD area within these crops is still small and this unmatchable OD area led to the decreased performance on the GoDARTS internal verification for all comparison methods and FunduSegmenter. Therefore, we conducted a mini group of internal and external verification experiments for DUNet, TransUNet and FunduSegmenter using the original images of GoDARTS, namely removing the OD center cropping operation.

## 4. Results

Table 2 shows the results of the internal verification and their comparisons with the baseline methods. External verification results are summarized in Supplementary Table S1 to S5, and domain generalization ones in Table 3 to 5. Results of the Group 1 ablation study (main ablation study to investigate proposed module effectiveness) are presented in Table 6; those of Group 2 and 3 in Supplementary Table S6 and S7. Note that 'step' in 'Best step' denotes one mini-batch. Since external verification used the models trained in the internal verification experiments for testing all the other datasets, the best steps shown in external result tables correspond to the best steps of internal verification results (omitted to save space in Table 2). In all result tables, bold identifies best performance, and underlining the second-best one. Our main aim is to compare the performance of different models using different datasets, rather than the performance of each individual model across the datasets.

### 4.1. Internal Verification Results

Table 2 presents the results of internal verification experiments. Our method outperformed the baselines in terms of the average OD, OC and all values across all datasets, and achieved both the best (OD: 92.52%, OC: 87.25%, All: 90.51%) and second-best scores (OD: 92.24%, OC: 87.15%, All: 90.37%). We also obtained the best DSC for Drishti-GS OD (97.34% [97.03%, 97.61%]) and OC (90.92% [89.13%, 92.48%]), for RIM-ONE-r3 OC (84.82% [82.60%, 86.94%]) and for REFUGE OC (87.19% [86.50%, 87.87%]). Additionally, we achieved the second-best scores for GoDARTS OD (78.99% [72.99%, 84.14%]) and IDRiD OD (96.94% [96.35%, 97.39%]). Our results were only slightly lower, hence competitive, than the best results for RIM-ONE-r3 OD (95.82% [95.11%, 96.40%]) and REFUGE OD (93.61% [93.32%, 93.90%]).

| *Methods* | *Aug* | *Datasets* | | | | | | | | *Average* | | |
|---|---|---|---|---|---|---|---|---|---|---|---|---|
| | | GoDARTS | IDRiD | Drishti-GS | | RIM-ONE-r3 | | REFUGE | | | | |
| | | OD | OD | OD | OC | OD | OC | OD | OC | OD | OC | All |
| nnU-Net | - | 17.96 [13.70, 22.71] | **96.99 [96.30, 97.58]** | 96.89 [95.92, 97.64] | 90.43 [88.67, 92.02] | **96.45 [95.38, 97.15]** | 82.57 [79.10, 85.58] | **95.85 [95.62, 96.06]** | 86.12 [85.42, 86.81] | 80.83 | 86.37 | 82.91 |
| DUNet | None | 75.58 [71.25, 79.67] | 94.28 [90.93, 96.37] | 95.65 [94.59, 96.52] | 84.35 [81.21, 87.09] | 93.49 [92.17, 94.58] | 79.05 [76.20, 81.68] | 89.32 [88.71, 89.93] | 78.28 [77.55, 79.01] | 89.66 | 80.56 | 86.25 |
| | Spatial | **79.42 [73.73, 84.15]** | 96.09 [94.43, 97.14] | 96.41 [95.59, 97.10] | 87.46 [85.27, 89.50] | 95.31 [94.37, 96.08] | 80.34 [77.16, 83.35] | 93.63 [93.32, 93.93] | 80.90 [79.92, 81.87] | 92.17 | 82.90 | 88.70 |
| | Designed | 75.35 [70.38, 80.07] | 96.15 [94.88, 97.02] | 97.18 [96.79, 97.54] | 89.11 [87.00, 90.83] | 95.58 [94.71, 96.33] | 82.02 [79.20, 84.69] | 92.63 [92.21, 93.02] | 85.32 [84.49, 86.06] | 91.38 | 85.48 | 89.17 |
| TransUNet | None | 66.13 [60.23, 72.14] | 95.20 [92.99, 96.61] | 96.66 [96.03, 97.15] | 87.49 [84.49, 90.04] | 95.23 [94.36, 95.95] | 80.47 [77.41, 83.17] | 90.73 [90.20, 91.25] | 80.18 [79.01, 81.30] | 88.79 | 82.71 | 86.51 |
| | Spatial | 68.44 [63.11, 73.85] | 95.49 [93.01, 97.13] | 97.29 [96.96, 97.59] | 89.66 [87.84, 91.23] | 95.71 [94.87, 96.35] | 81.35 [78.35, 84.16] | 91.94 [91.41, 92.45] | 83.38 [82.48, 84.28] | 89.77 | 84.80 | 87.91 |
| | Designed | 65.05 [58.06, 71.97] | 96.39 [95.56, 97.07] | 97.07 [96.23, 97.58] | 88.29 [85.51, 90.61] | 96.06 [95.40, 96.61] | 82.76 [79.85, 85.36] | 93.76 [93.36, 94.14] | 82.29 [81.27, 83.27] | 89.67 | 84.45 | 87.71 |
| Ours | None | 73.47 [64.34, 81.31] | 95.91 [95.06, 96.59] | 96.91 [96.55, 97.25] | 88.24 [85.90, 90.36] | 95.20 [94.42, 95.88] | 81.95 [79.48, 84.25] | 92.94 [92.61, 93.26] | 84.26 [83.53, 84.98] | 90.89 | 84.82 | 88.61 |
| | Spatial | 78.26 [71.16, 84.25] | 96.94 [96.35, 97.39] | 97.25 [96.94, 97.55] | **90.92 [89.13, 92.48]** | 95.82 [95.11, 96.40] | 83.63 [81.14, 85.97] | 92.91 [92.59, 93.22] | **87.19 [86.50, 87.87]** | 92.24 | **87.25** | 90.37 |
| | Designed | 78.99 [72.99, 84.14] | 96.88 [96.31, 97.33] | **97.34 [97.03, 97.61]** | 90.85 [88.87, 92.64] | 95.78 [94.93, 96.46] | **84.82 [82.60, 86.94]** | 93.61 [93.32, 93.90] | 85.78 [85.07, 86.49] | **92.52** | 87.15 | **90.51** |

*Table 2. Results of our method compared with state-of-the-art baselines on the internal verification task. All figures are DSC (%) with 95% CI. Average OD, OC and 'All' denote the simple average of OD, OC and all DSC.*

### 4.2. External Verification Results

Supplementary Table S1 to S5 present the results of external verification experiments. Our method outperformed the baselines in terms of the average OD and all values when using each of all datasets as the source dataset, achieving both the best and the second-best scores. Our results also achieved the best scores in terms of the average OC values when using Drishti-GS and RIM-ONE-r3 as source datasets. We achieved the second-best score only for the average OC values with REFUGE as the source dataset.

### 4.3. Domain Generalization Results

Table 3 to 5 present the results of domain generalization experiments. Our method outperformed the state-of-the-art task-specific baselines, DoFE and RAM-DSIR, in terms of average performance, but was slightly lower than TVConv. Note that the ground truth of Domain 1 (Drishti-GS) was generated by the agreement of 3 experts out of 4 (Section 3.2.2), rather than 4 out of 4 as used in DoFE, RAM-DSIR, and TVConv. Consequently, our absolute DSC values are not directly comparable to those reported in their original papers.

| **Metric: DSC (%)** | *Target domain* | | | | | | | | *Average* | | |
|---|---|---|---|---|---|---|---|---|---|---|---|
| *Methods* | Domain 1 | | Domain 2 | | Domain 3 | | Domain 4 | | | | |
| | OD | OC | OD | OC | OD | OC | OD | OC | OD | OC | All |
| DoFE | 92.93 | 74.31 | 87.30 | 73.42 | 93.27 | **86.40** | 93.38 | 86.97 | 91.72 | 80.28 | 86.00 |
| RAM-DSIR | 95.24 | <u>75.85</u> | 86.51 | <u>74.32</u> | **95.36** | 85.74 | <u>94.54</u> | 83.02 | 92.91 | 79.73 | 86.32 |
| TVConv | <u>95.74</u> | **84.12** | **89.36** | **76.55** | <u>95.11</u> | 86.11 | 93.69 | 84.08 | **93.48** | **82.72** | **88.10** |
| Ours (None) | 94.62 | 73.27 | 88.03 | 74.28 | 92.88 | 86.05 | **94.88** | 85.56 | 92.60 | 79.79 | 86.20 |
| Ours (Spatial) | **95.78** | 74.49 | <u>88.55</u> | 72.54 | 94.16 | <u>86.33</u> | 93.54 | <u>87.86</u> | <u>93.01</u> | <u>80.31</u> | <u>86.66</u> |
| Ours (Designed) | 95.36 | 71.81 | 87.32 | 73.57 | 93.81 | 86.27 | 93.88 | **88.11** | 92.59 | 79.94 | 86.27 |

*Table 3. Results of our method compared with state-of-the-art domain generalization segmentation networks. All figures are DSC (%). Average OD, OC and 'All' denote the simple average of OD, OC and all DSC.*

| **Metric: HD95** | *Target domain* | | | | | | | | *Average* | | |
|---|---|---|---|---|---|---|---|---|---|---|---|
| *Methods* | Domain 1 | | Domain 2 | | Domain 3 | | Domain 4 | | | | |
| | OD | OC | OD | OC | OD | OC | OD | OC | OD | OC | All |
| DoFE | 21.90 | 46.75 | 33.18 | 33.02 | 19.03 | 19.53 | 14.02 | 15.64 | 22.03 | 28.74 | 25.38 |
| RAM-DSIR | 18.76 | 49.55 | 32.70 | 36.52 | **14.28** | 20.36 | 14.70 | 25.36 | 20.11 | 32.95 | 26.53 |
| TVConv | 15.81 | **34.09** | 27.86 | **31.40** | 14.68 | 20.13 | 15.10 | 18.14 | **18.36** | **25.94** | **22.15** |
| Ours (None) | 18.96 | 56.55 | 34.66 | 35.62 | 21.85 | 22.33 | **13.25** | 14.15 | 22.18 | 32.16 | 27.17 |
| Ours (Spatial) | 16.65 | 47.37 | **24.94** | 33.74 | 16.96 | 20.15 | 16.70 | 13.96 | 18.81 | 28.81 | 23.81 |
| Ours (Designed) | **15.45** | 50.45 | 27.08 | 32.95 | 17.72 | **19.30** | 16.56 | **13.18** | 19.20 | 28.97 | 24.09 |

*Table 4. Results of our method compared with state-of-the-art domain generalization segmentation networks. All figures are HD95 values. Average OD, OC and 'All' denote the simple average of OD, OC and all HD95 values.*

| **Metric: ASD** | *Target domain* | | | | | | | | *Average* | | |
|---|---|---|---|---|---|---|---|---|---|---|---|
| *Methods* | Domain 1 | | Domain 2 | | Domain 3 | | Domain 4 | | | | |
| | OD | OC | OD | OC | OD | OC | OD | OC | OD | OC | All |
| DoFE | 12.38 | 28.41 | 19.61 | 19.81 | 9.56 | **9.70** | 7.18 | 7.54 | 12.18 | 16.37 | 14.27 |
| RAM-DSIR | 8.63 | 27.09 | 20.21 | 18.72 | **6.70** | 10.03 | 6.21 | 15.28 | 10.44 | 17.78 | 14.11 |
| TVConv | **7.56** | **18.43** | 17.02 | **16.88** | 7.01 | 10.81 | 6.94 | 8.89 | **9.63** | **13.75** | **11.69** |
| Ours (None) | 9.47 | 30.53 | 18.83 | 19.62 | 9.98 | 10.47 | **5.74** | 7.25 | 11.01 | 16.97 | 13.99 |
| Ours (Spatial) | 7.61 | 29.04 | **16.89** | 21.11 | 8.36 | 9.89 | 7.45 | 6.81 | 10.08 | 16.71 | 13.40 |
| Ours (Designed) | 8.18 | 31.60 | 18.83 | 20.40 | 8.77 | 9.72 | 7.05 | **6.54** | 10.71 | 17.07 | 13.89 |

*Table 5. Results of our method compared with state-of-the-art domain generalization segmentation networks. All figures are ASD values. Average OD, OC and 'All' denote the simple average of OD, OC and all ASD values.*

### 4.4. Ablation Study Results

Table 6 presents the results of Group 1 ablation study experiments; Group 2 and 3 results are in Supplementary Table S6 and S7. The results of Group 1 and Group 2 indicate the effectiveness of each module added to the model. Both performance and convergence speed improved significantly with the addition of the proposed modules. The results of Group 3 show that our method performed even better (92.75% to 93.20%) on the external verification task without applying OD center cropping, while the performance of DUNet (87.62% to 73.25%) and TransUNet (87.80% to 85.33%) decreased significantly.

| *No.* | *Module* | *Model* | *Size* | *Best step* | *Datasets* | | | | | | | | *Average* |
|---|---|---|---|---|---|---|---|---|---|---|---|---|---|
| | | | | | Source | | Target | | | | | | |
| | | | | | REFUGE | | GoDA RTS | IDRiD | Drishti-GS | | RIM-ONE-r3 | | |
| | | | | | OD | OC | OD | OD | OD | OC | OD | OC | All |
| 1 | Decoder | RET + Seg(6) | $224^2$ | 37,956 | 89.91 | 78.95 | 80.87 | 93.09 | 94.38 | 70.09 | 89.38 | 73.26 | 83.74 |
| 2 | | RET + Seg(4) | $224^2$ | 23,220 | 92.65 | 82.36 | 84.50 | 94.76 | 96.13 | 63.66 | 86.09 | 73.71 | 84.23 |
| 3 | | RET(2) + Seg(2) | $224^2$ | 8,640 | 91.54 | 81.56 | 80.31 | 94.25 | 95.65 | 69.88 | 88.35 | 80.88 | 85.30 |
| 4 | | **RET + Seg(2)** | $224^2$ | 17,160 | 93.09 | 83.33 | 83.95 | 93.88 | 94.72 | 71.32 | 83.07 | 78.10 | **85.18** |
| 5 | Post-adapter | **Best + Post-A** | $224^2$ | 888 | 93.39 | 85.65 | 87.10 | 95.63 | 96.41 | 71.30 | 87.55 | 81.98 | **87.38** |
| 6 | Pre-adapter | Pre-A + Best | $224^2$ | 2,868 | 93.06 | 86.67 | 83.93 | 95.70 | 95.95 | 77.70 | 87.29 | 81.73 | 87.76 |
| 7 | | Best | $464^2$ | 372 | 94.35 | 84.31 | 80.76 | 94.16 | 95.44 | 52.29 | 87.97 | 68.40 | 82.21 |
| 8 | | Pre-A + Best | $464^2$ | 5,928 | 92.58 | 85.61 | 80.00 | 95.97 | 95.73 | 74.00 | 86.34 | 79.21 | 86.18 |
| 9 | | Best (Tune pos) | $320^2$ | 40 | 93.82 | 84.96 | 85.26 | 94.90 | 96.47 | 73.82 | 85.34 | 80.99 | 86.94 |
| 10 | | Pre-A + Best | $320^2$ | 228 | 93.65 | 86.61 | 86.88 | 95.85 | 96.56 | 74.92 | 84.57 | 77.98 | 87.13 |
| 11 | | Pre-A + Best (Tune RET(2)) | $224^2$ | 972 | 93.65 | 85.78 | 86.32 | 96.24 | 96.83 | 67.52 | 87.23 | 80.11 | 86.71 |
| 12 | | **Pre-A + Best** | $256^2$ | 1,764 | 93.45 | 86.01 | 83.43 | 95.76 | 95.74 | 77.16 | 86.87 | 79.35 | **87.22** |
| | | | | | Source | | Target | | | | | | |
| | | | | | Drishti-GS | | GoDA RTS | IDRiD | RIM-ONE-r3 | | REFUGE | | |
| | | | | | OD | OC | OD | OD | OD | OC | OD | OC | All |
| 13 | Skip | Best | $256^2$ | 8,598 | 97.28 | 89.93 | 90.90 | 95.71 | 86.80 | 71.60 | 62.74 | 46.70 | 80.21 |
| 14 | | **Best + Skip** | $256^2$ | 5,033 | 97.55 | 90.74 | 89.48 | 95.93 | 84.11 | 66.37 | 80.97 | 71.59 | **84.59** |
| 15 | ViT block adapter | Best + A between LN and MLP | $256^2$ | 296 | 96.86 | 89.88 | 88.00 | 95.91 | 87.62 | 64.19 | 85.48 | 63.52 | 83.93 |
| 16 | | **FunduSegmenter** | $256^2$ | 1,250 | 97.34 | 90.85 | 88.96 | 96.13 | 85.59 | 65.61 | 88.63 | 73.61 | **85.84** |

*Table 6. Results of ablation study Group 1 experiments. All figures are DSC (%). Average 'All' denotes the simple average of all DSC. 'RET' denotes RETFound; 'RET(number)' unfreezing last 'number' layers of RETFound; 'Seg' Segmenter decoder; 'Seg(number)' the Segmenter decoder with 'number' layers; 'A' adapter; 'pos' position embeddings; 'Best' the final selection of each module.*

## 5. Discussion

### 5.1. Model Stability and Generalization Capability

As shown in Table 2, our method outperformed the baselines in average DSC results. nnU-Net achieved competitive results and obtained the best results on IDRiD, RIM-ONE-r3 and REFUGE OD segmentation. We attribute this to the careful design of its pre-processing approach. The success of nnU-Net represents the pinnacle of *dedicated* networks for *specific* tasks and datasets. However, the performance of nnU-Net decreased rapidly when generalizing to unseen out-of-distribution data. The GoDARTS dataset was selected randomly from a large cohort, so the distributions of the training and testing datasets were quite different. In this case, nnU-Net only achieved 17.96% DSC. By contrast, our method did not show weaknesses on *any* datasets used in our experiments. This is likely due to the strong general representations from the frozen RETFound encoder and the effective adaptation to the segmentation task by our proposed modules.

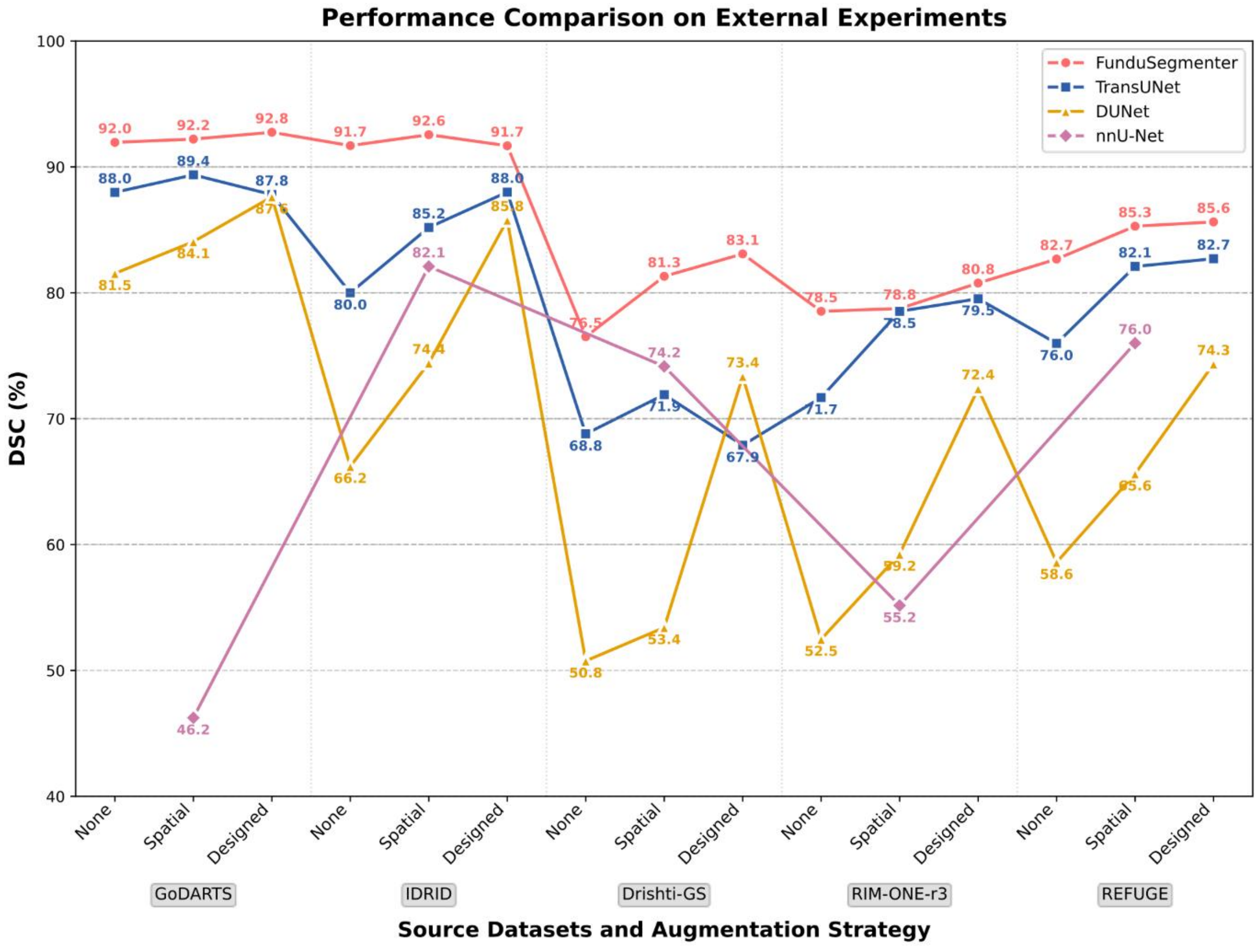

*Figure 4. Average DSC results of nnU-Net, DUNet, TransUNet and our FunduSegmenter when using each dataset as the source dataset. Note that nnU-Net followed the original design, so the augmentation strategy shown in this figure should be ignored for it.*

Furthermore, our method demonstrated strong generalization capability in external verification experiments. Figure 4 presents the average DSC results of nnU-Net, DUNet, TransUNet and our method when using each dataset in turn as the source domain (a summary of internal and external verification results). Note that values are lower for Drishti-GS, RIM-ONE-r3 and REFUGE since OC results are included. Our method clearly outperformed all baselines, exhibiting stability and being only slightly affected by the changes in data scarcity. For OD-only segmentation, DSC remains about 92% regardless of the source data or augmentation strategy. For joint OD and OC segmentation, the performance was affected by the changes in data scarcity but still remained ~83% DSC. By contrast, the performance of all baseline models varied drastically with changes in the distribution of the source domain and the scarcity of data.

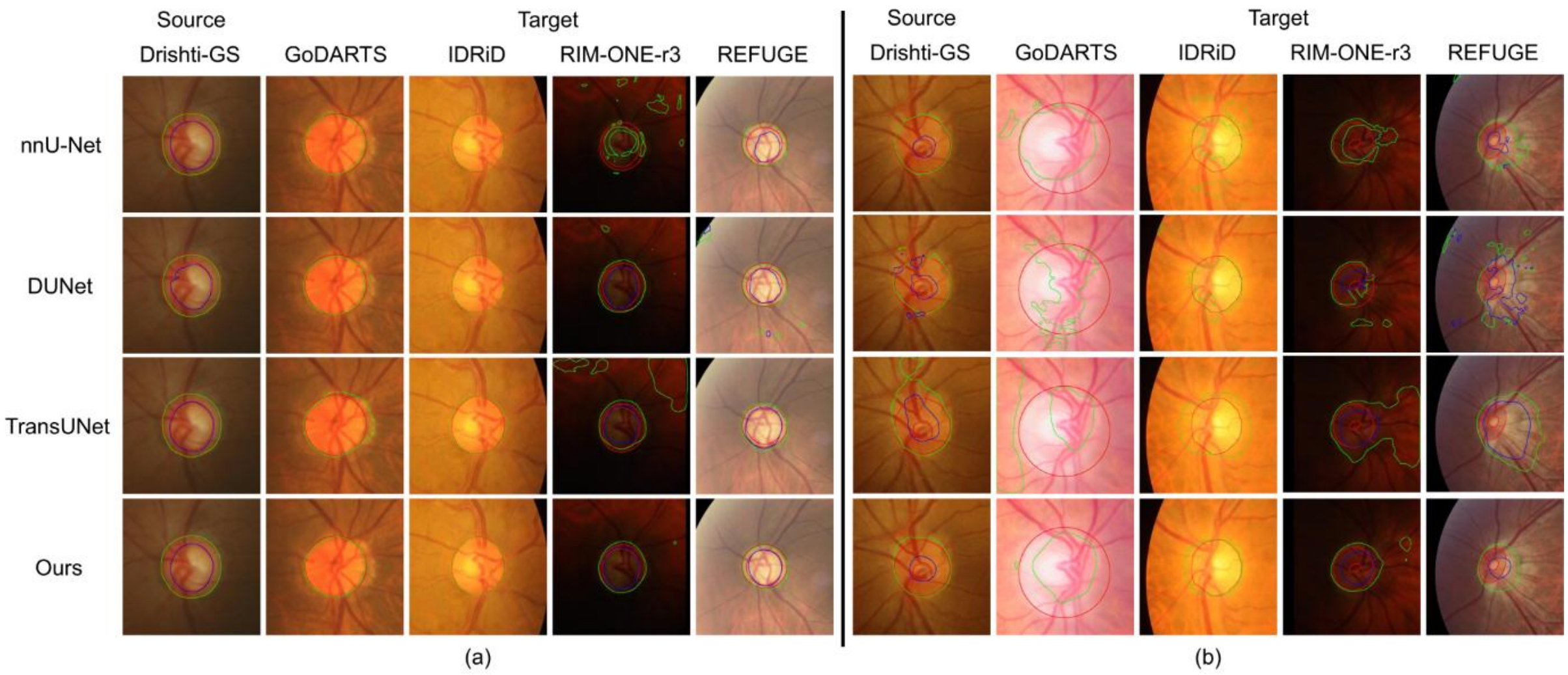

*Figure 5. Visual examples of segmentation results of nnU-Net, DUNet, TransUNet and our FunduSegmenter when using Drishti-GS as the source dataset. Designed augmentations were applied. (a) The best result of FunduSegmenter and the results of other baselines on the same images. (b) The worst result of FunduSegmenter and the results of other baselines on the same images.*

Figure 5 indicates the best result and worst results of our method when using Drishti-GS (the smallest dataset: 50 training images) as the source domain and the results of baselines on the same images. Our method's best results in both internal and external verification almost match the ground truth. The worst results involve only oversegmentation or undersegmentation in the correct positions. Completely wrong segmentations are confined to a few small areas and can be easily removed by a post-processing method that preserves the largest connected component. By contrast, all baselines produced many completely wrong segmentations which cannot be corrected by post-processing approaches.

## 5.2. Effectiveness, Potential and Generalizability of Proposed Modules

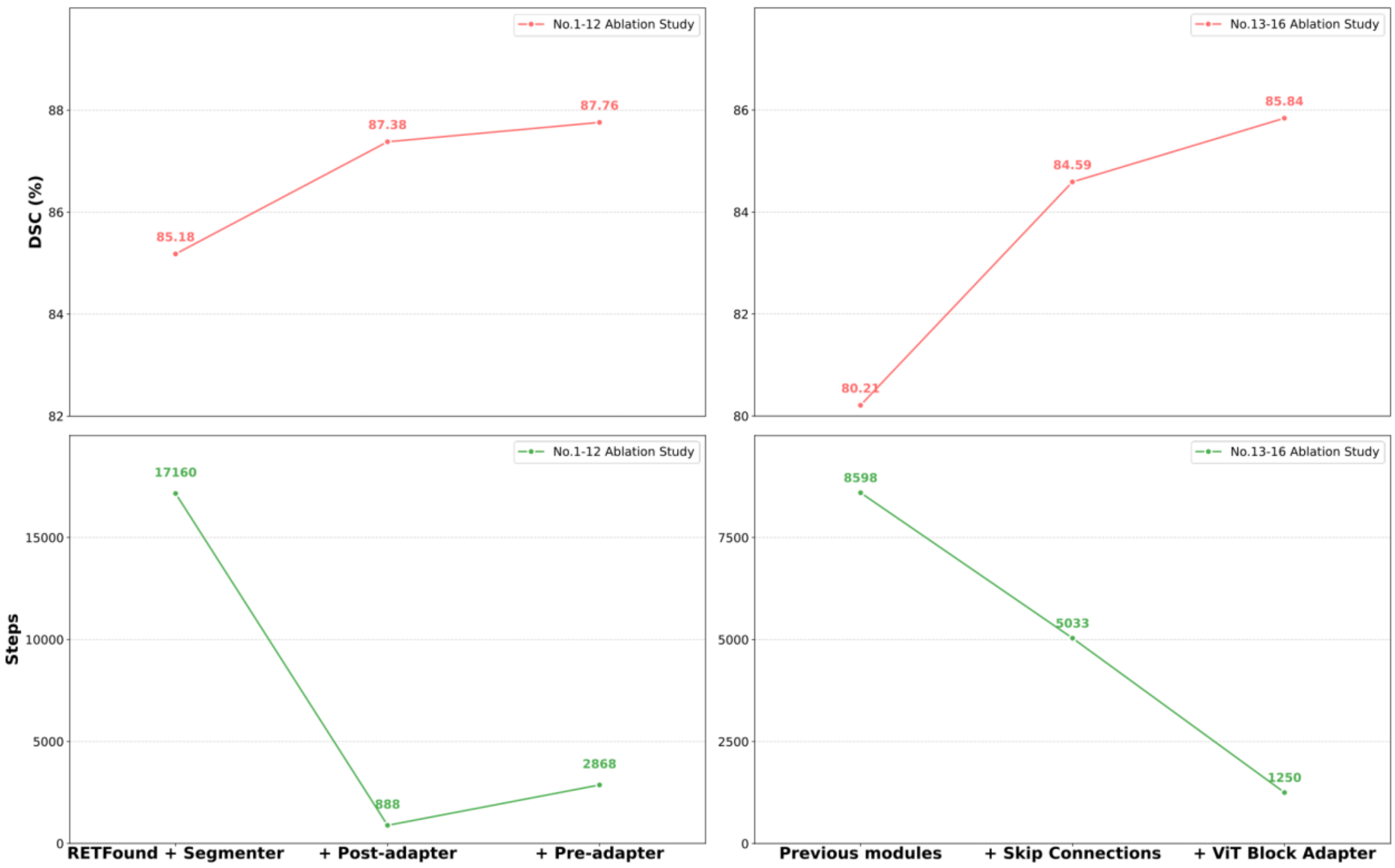

*Figure 6. Summary of ablation study results for the proposed modules. Upper left: DSC scores of RETFound with incremental integration of the Segmenter decoder, Post-adapter and Pre-adapter (key configurations in No. 1-12 ablation study which were trained on REFUGE). Upper right: DSC scores of the model with further incremental integration of skip connections and ViT Block Adapter (key configurations in No. 13-16 which were trained on Drishti-GS). Bottom left: training steps at which the highest DSC scores were achieved (No. 1-12 ablation study). Bottom right: same for No. 13-16.*

Our proposed modules significantly improved both performance and convergence speed (Table 6). Figure 6 summarizes the ablation study results for the proposed modules with key configurations. We also tried unfreezing the last two layers of RETFound (No.4 ablation study). Although this performed slightly better for the ‘RETFound + Segmenter decoder’ architecture, its performance decreased with the addition of other modules (No.11 ablation study), leading us to abandon this option. By comparing between No.7 and No.8, and between No.9 and No.10 ablation study, our Pre-adapter outperformed the position interpolation method. Furthermore, our Pre-adapter keeps the training time per epoch at about 11 seconds with our experiment pipeline and supports up to 592 × 592 input size (without skip connections and ViT block adapter). By contrast, position interpolation without fine-tuning only supports a maximum input size of 464 × 464 and 320 × 320 with fine-tuning, and increases the training time of one epoch to about 32 seconds and 18 seconds, respectively. As expected, the performance of the model with Pre-adapter slightly decreased with the increasing input size. The final choice of 256 × 256 was for fair comparison with baselines, although 224 × 224 achieved the best performance. The addition of skip connections with CBAM and ViT block adapter brought significant improvements in training stability, convergence speed, and performance. Supplementary Figure S2 presents the training progression for different model configurations on IDRiD (more details in Supplementary Table S6). When using only Segmenter decoder for adaptation, the training loss remained at about 0.9 for 50,000 steps, and then dropped to and remained at about 0.15 for 70,000 steps before converging abruptly. This unstable training was more pronounced when the training dataset was small and the task was OD-only segmentation. We attribute this to the difficulty of reconstructing spatial maps from sequence representations (features from RETFound) under data scarcity and limited supervised signal. Our proposed Post-adapter, skip connections with CBAM and ViT block adapter effectively addressed this problem through progressively reconstructing spatial maps from sequences, introducing multi-scale information and providing better adaptation to the segmentation task, respectively.

As shown in Supplementary Table S7, although DUNet, TransUNet and our method all showed improved performance on the GoDARTS internal verification when the original images were used instead of OD center cropped images, OD center cropping significantly benefited DUNet and TransUNet in external verification. By contrast, our method performed even better in external verification without OD center cropping. This indicates that OD center cropping, while a common and generally beneficial pre-processing step for OD and OC segmentation (with the GoDARTS internal verification being a notable exception), may not be optimal for our FunduSegmenter. Our focus was on model architecture, and we did not extensively optimize dataset processing. This suggests there is further potential for improvement with our model.

On both generalization tasks (external verification and domain generalization), our method was only slightly behind TVConv. However, TVConv benefits from strong prior positional knowledge and may degrade when facing images with severe layout transformations [21]. This means it can only be applied to targets with a certain layout (e.g., OD/OC processed by OD center cropping), but cannot be extended to other complex segmentation tasks like fundus vessel segmentation (the layout of tiny vessels is complex) [1] or glioma segmentation (the locations vary) [46]. By contrast, we did not use any prior knowledge in the proposed modules, so our proposed modules may serve as a general adaptation method for any other FMs with ViT as the encoder and trained by SSL.

### 5.3. Limitations and Future Work

Although our method has shown promising performance, it still has at least two weaknesses. First, its results were slightly behind those of TVConv, even though the latter is task-specific and has restrictions. Second, our method was not strong enough for OC segmentation, unlike outperforming all baselines for OD segmentation. Specifically, our results trailed behind those of TVConv, primarily in OC segmentation. Also, we did not achieve the best OC results when using REFUGE as the source dataset. Although these can be regarded as arguably minor limitations considering overall performance, we plan to investigate methods to further improve our model.

As discussed in Section 5.2, our method has strong potential and generalization. Further work includes designing a standard data processing pipeline and expanding the application of our model to other important tasks in retinal image analysis, such as vessel segmentation and characterization, and the discovery of biomarkers for systemic conditions [47-49]. Additionally, we also plan to apply our model to other image domains.

## 6. Conclusion

In this paper we proposed FunduSegmenter, to our best knowledge, the first adaptation of the RETFound foundation model to retinal image segmentation. RETFound and related work [16-18] focused on classification. We proposed a novel architecture to explore the potential of the general representation learned by RETFound for a fundamental task in retinal image analysis, namely segmenting the OD and OC in fundus camera images.

Our FunduSegmenter achieved high performance, nearly always surpassing that of six state-of-the-art baseline methods in all internal verification, external verification and domain generalization experiments. The model maintained consistently high performance across all experiments, regardless of the dataset or augmentation strategy used, indicating strong stability and generalization capability. We hope our comprehensive experiments can serve as a benchmark for joint OD and OC segmentation on cross-distribution datasets and for adapting SSL FMs to segmentation tasks, as there is currently very little work on this topic. Furthermore, our proposed modules are general and can be extended to the fine-tuning of other FMs with ViT as the encoder trained through SSL.

## Acknowledgments

The authors acknowledge Stephen Hogg (VAMPIRE group, Dundee) for useful information regarding code and papers.

## Supplementary Materials

## Supplementary Figures

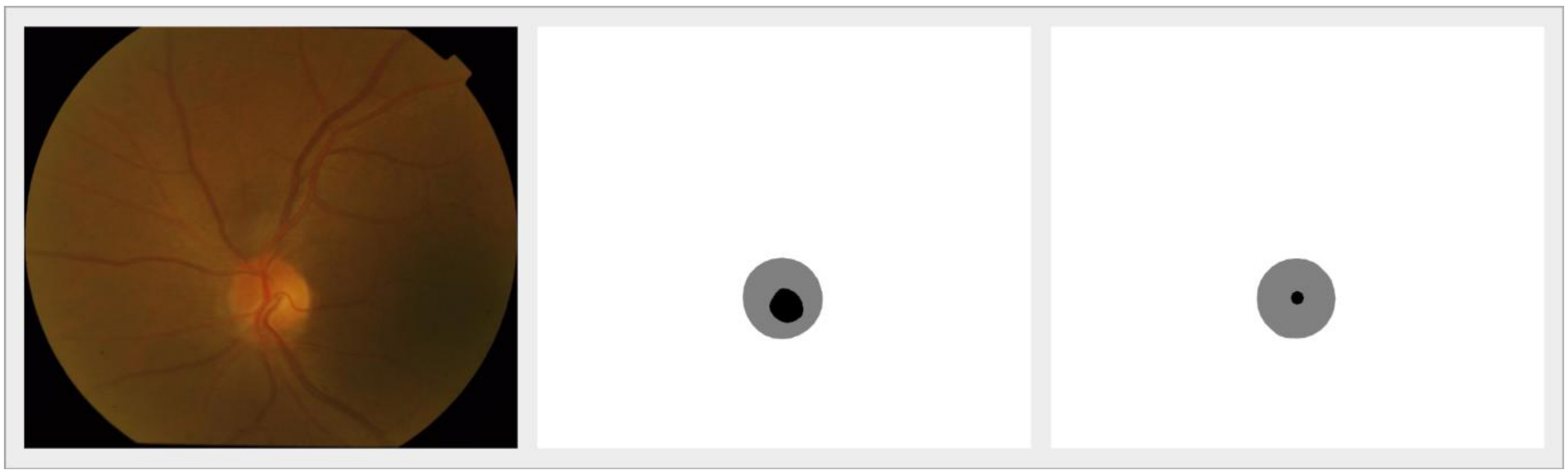

Figure S1. Drishti-GS segmentation mask example (drishtiGS_096). Left: Image. Middle: segmentation mask produced by the agreement from three experts (our way). Right: segmentation mask produced by the agreement from all experts (DoFE's way).

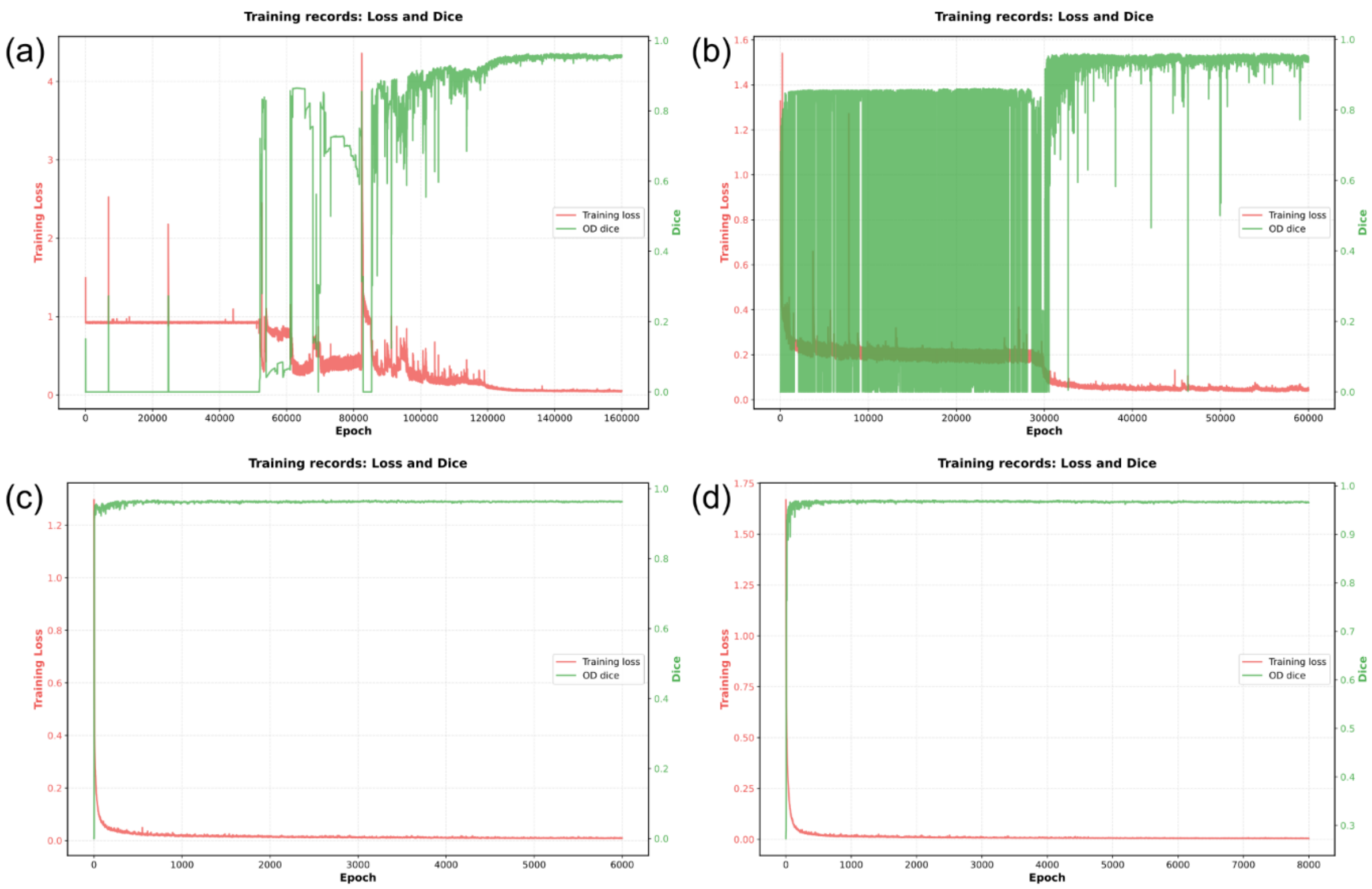

Figure S2. Training progression of model architectures with key proposed modules on IDRiD. (a) RETFound + Segmenter. (b) Pre-adapter + (a) + Post-adapter. (c) Skip connections with CBAM + (b). (d) FunduSegmenter ((c) + ViT block adapter).

## Supplementary Tables

| *Methods* | *Aug* | *Best step* | *Datasets* | | | | *Average* |
|---|---|---|---|---|---|---|---|
| | | | Target (Source: GoDARTS) | | | | |
| | | | IDRiD | Drishti-GS | RIM-ONE-r3 | REFUGE | |
| | | | OD | OD | OD | OD | OD |
| nnU-Net | - | - | 9.88 [7.70, 12.24] | 35.40 [31.85, 39.06] | 52.47 [47.93, 56.82] | 87.19 [86.14, 88.19] | 46.24 |
| DUNet | None | 80,070 | 93.16 [91.15, 94.77] | 82.18 [76.63, 87.00] | 59.85 [51.48, 67.54] | 90.98 [90.43, 91.50] | 81.54 |
| | Spatial | 13,480 | 95.09 [93.97, 96.06] | 84.11 [79.49, 88.28] | 66.08 [58.18, 73.09] | 90.95 [90.40, 91.46] | 84.06 |
| | Designed | 3,595 | 94.12 [92.40, 95.47] | 91.72 [89.44, 93.41] | 73.84 [67.34, 79.56] | 90.79 [90.24, 91.31] | 87.62 |
| TransUNet | None | 1,810 | 88.09 [85.98, 90.01] | 81.78 [79.28, 84.13] | 90.02 [88.53, 91.34] | **92.04 [91.48, 92.57]** | 87.98 |
| | Spatial | 3,110 | 91.37 [90.06, 92.62] | 83.65 [81.24, 85.91] | **91.00 [90.01, 91.93]** | 91.45 [90.88, 91.98] | 89.37 |
| | Designed | 920 | 88.59 [85.33, 91.32] | 82.17 [79.30, 84.77] | 89.22 [88.11, 90.29] | 91.21 [90.62, 91.78] | 87.80 |
| Ours | None | 155 | 92.12 [91.24, 92.93] | 94.95 [94.18, 95.60] | 88.95 [88.06, 89.78] | 91.79 [91.45, 92.12] | 91.95 |
| | Spatial | 620 | 92.99 [92.20, 93.74] | 95.30 [94.43, 96.07] | 89.90 [89.12, 90.66] | 90.63 [90.17, 91.07] | 92.21 |
| | Designed | 910 | **95.41 [94.91, 95.92]** | **95.54 [94.83, 96.18]** | 88.73 [87.93, 89.51] | 91.31 [90.85, 91.74] | **92.75** |

Table S1. Results of our method compared with state-of-the-art baselines on the external verification task (Source: GoDARTS). All figures are DSC (%) with 95% CI. Average OD denotes the simple average of OD DSC.

| *Methods* | *Aug* | *Best step* | *Datasets* | | | | *Average* |
|---|---|---|---|---|---|---|---|
| | | | Target (Source: IDRiD) | | | | |
| | | | GoDARTS | Drishti-GS | RIM-ONE-r3 | REFUGE | |
| | | | OD | OD | OD | OD | OD |
| nnU-Net | - | - | **93.11 [91.63, 94.56]** | 92.97 [89.01, 95.57] | 79.07 [72.95, 84.14] | 63.21 [59.33, 66.99] | 82.09 |
| DUNet | None | 1,775 | 78.11 [69.77, 85.01] | 69.21 [60.87, 76.90] | 40.47 [32.39, 48.26] | 77.04 [75.82, 78.23] | 66.21 |
| | Spatial | 16,756 | 82.01 [75.48, 87.78] | 76.48 [69.16, 83.10] | 47.96 [39.42, 56.15] | 91.24 [90.53, 91.89] | 74.42 |
| | Designed | 1,046 | 89.35 [87.00, 91.65] | 93.25 [91.92, 94.52] | 76.03 [71.82, 79.88] | 84.49 [83.55, 85.42] | 85.78 |
| TransUNet | None | 6,220 | 71.96 [60.56, 82.20] | 95.87 [95.09, 96.52] | 81.93 [78.70, 84.73] | 70.25 [68.87, 71.65] | 80.00 |
| | Spatial | 897 | 78.29 [69.58, 85.97] | 95.13 [94.17, 96.01] | 80.84 [77.62, 83.67] | 86.49 [85.50, 87.46] | 85.19 |
| | Designed | 3,550 | 87.55 [84.55, 90.04] | 94.66 [93.83, 95.46] | 83.02 [80.28, 85.27] | 86.75 [85.68, 87.80] | 88.00 |
| Ours | None | 93 | 89.97 [87.84, 91.94] | 96.40 [95.78, 96.90] | **89.61 [87.74, 91.06]** | 90.79 [90.32, 91.24] | 91.69 |
| | Spatial | 5,242 | 91.25 [89.59, 92.85] | **97.09 [96.79, 97.37]** | 88.17 [87.11, 89.22] | **93.74 [93.33, 94.13]** | **92.56** |
| | Designed | 1,845 | 90.85 [88.97, 92.54] | 97.05 [96.75, 97.34] | 86.91 [85.96, 87.81] | 91.90 [91.32, 92.46] | 91.68 |

Table S2. Results of our method compared with state-of-the-art baselines on the external verification task (Source: IDRiD). All figures are DSC (%) with 95% CI. Average OD denotes the simple average of OD DSC.

| *Methods* | *Aug* | *Best step* | *Datasets* | | | | | | *Average* | | |
|---|---|---|---|---|---|---|---|---|---|---|---|
| | | | Target (Source: Drishti-GS) | | | | | | | | |
| | | | GoDARTS | IDRiD | RIM-ONE-r3 | | REFUGE | | | | |
| | | | OD | OD | OD | OC | OD | OC | OD | OC | All |
| nnU-Net | - | - | 69.76 [63.82, 75.81] | 66.63 [62.48, 70.59] | 84.10 [81.59, 86.32] | 60.19 [54.81, 65.33] | 88.16 [87.07, 89.23] | **76.03 [74.51, 77.55]** | 77.16 | 68.11 | 74.15 |
| DUNet | None | 8,668 | 66.18 [57.40, 74.67] | 68.83 [61.98, 75.35] | 74.90 [69.57, 79.42] | 54.76 [49.43, 60.04] | 28.39 [27.43, 29.41] | 11.52 [10.73, 12.38] | 59.58 | 33.14 | 50.76 |
| | Spatial | 10,764 | 65.26 [56.86, 73.82] | 78.45 [70.90, 85.43] | 76.20 [70.79, 80.95] | 56.34 [50.70, 61.86] | 32.18 [30.74, 33.76] | 12.05 [10.96, 13.20] | 63.02 | 34.20 | 53.41 |
| | Designed | 7,703 | 80.30 [75.68, 84.85] | 94.03 [92.68, 95.28] | 85.54 [83.73, 87.17] | **69.49 [64.77, 73.86]** | 67.39 [66.15, 68.61] | 43.43 [41.56, 45.34] | 81.82 | 56.46 | 73.36 |
| TransUNet | None | 413 | 69.06 [57.62, 79.80] | 93.47 [91.33, 95.09] | 77.21 [74.28, 79.90] | 53.60 [48.19, 59.03] | 66.58 [65.40, 67.74] | 52.79 [51.03, 54.54] | 76.58 | 53.20 | 68.79 |
| | Spatial | 3,979 | 81.21 [75.76, 85.97] | 94.27 [92.25, 95.85] | 82.60 [80.67, 84.32] | 63.47 [58.42, 68.29] | 66.39 [65.40, 67.37] | 43.51 [41.97, 45.02] | 81.12 | 53.49 | 71.91 |
| | Designed | 4,740 | 79.82 [72.57, 86.35] | 93.36 [90.56, 95.62] | 80.69 [78.31, 82.85] | 58.85 [53.68, 63.81] | 60.52 [59.30, 61.74] | 34.07 [32.70, 35.44] | 78.60 | 46.46 | 67.89 |
| Ours | None | 107 | 88.14 [85.19, 90.69] | 90.81 [85.57, 95.10] | 80.23 [76.45, 83.33] | 55.10 [49.43, 60.54] | 85.72 [84.81, 86.59] | 59.05 [57.42, 60.63] | 86.23 | 57.08 | 76.51 |
| | Spatial | 5,181 | 83.44 [78.09, 88.43] | 95.78 [94.37, 96.71] | **85.76 [84.45, 87.07]** | 67.74 [63.25, 72.01] | **90.07 [89.46, 90.65]** | 65.04 [63.75, 66.38] | 88.76 | 66.39 | 81.31 |
| | Designed | 1,250 | **88.96 [86.23, 91.34]** | **96.13 [95.44, 96.68]** | 85.59 [84.54, 86.61] | 65.61 [61.15, 69.90] | 88.63 [88.07, 89.17] | 73.61 [72.55, 74.67] | **89.83** | **69.61** | **83.09** |

Table S3. Results of our method compared with state-of-the-art baselines on the external verification task (Source: Drishti-GS). All figures are DSC (%) with 95% CI. Average OD, OC and 'All' denote the simple average of OD, OC and all DSC.

| *Methods* | *Aug* | *Best step* | *Datasets* | | | | | | *Average* | | |
|---|---|---|---|---|---|---|---|---|---|---|---|
| | | | Target (Source: RIM-ONE-r3) | | | | | | | | |
| | | | GoDARTS | IDRiD | Drishti-GS | | REFUGE | | | | |
| | | | OD | OD | OD | OC | OD | OC | OD | OC | All |
| nnU-Net | - | - | 55.01 [50.71, 58.99] | 54.50 [51.30, 57.66] | 82.90 [80.19, 85.44] | 58.45 [54.00, 62.77] | 51.40 [50.26, 52.53] | 28.70 [27.34, 30.14] | 60.95 | 43.58 | 55.16 |
| DUNet | None | 1,862 | 46.93 [39.52, 54.43] | 60.30 [55.30, 65.51] | 81.59 [78.39, 84.41] | 60.44 [56.08, 64.50] | 36.04 [35.13, 37.01] | 29.59 [28.32, 30.86] | 56.22 | 45.02 | 52.48 |
| | Spatial | 2,448 | 50.75 [43.40, 58.49] | 65.93 [60.32, 71.13] | 78.72 [74.76, 82.29] | 60.79 [56.93, 64.51] | 46.92 [45.69, 48.20] | 52.32 [50.43, 54.26] | 60.58 | 56.56 | 59.24 |
| | Designed | 2,600 | 68.63 [64.00, 73.21] | 68.36 [61.72, 75.07] | 89.37 [87.77, 90.73] | 71.69 [68.72, 74.47] | 71.38 [70.40, 72.36] | 64.92 [63.30, 66.53] | 74.44 | 68.31 | 72.39 |
| TransUNet | None | 1,292 | 68.07 [62.80, 72.78] | 76.83 [72.52, 80.71] | 88.79 [87.96, 89.54] | **72.15 [69.58, 74.66]** | 75.74 [74.66, 76.78] | 48.47 [46.19, 50.81] | 77.36 | 60.31 | 71.68 |
| | Spatial | 478 | 71.36 [66.42, 75.84] | 86.21 [84.91, 87.43] | 90.55 [89.89, 91.15] | 64.82 [61.36, 68.17] | 87.05 [86.67, 87.43] | 71.21 [69.77, 72.63] | 83.79 | 68.02 | 78.53 |
| | Designed | 622 | 73.84 [69.89, 77.41] | 88.62 [87.56, 89.58] | 91.24 [90.58, 91.87] | 69.70 [66.65, 72.64] | 87.40 [87.00, 87.79] | 66.38 [64.43, 68.25] | 85.28 | 68.04 | 79.53 |
| Ours | None | 398 | 73.12 [69.29, 76.58] | 84.04 [82.77, 85.27] | 91.39 [90.75, 91.97] | 61.30 [57.78, 64.85] | 88.02 [87.63, 88.40] | 73.29 [72.20, 74.39] | 84.14 | 67.30 | 78.53 |
| | Spatial | 3,968 | 69.63 [65.39, 73.66] | 85.37 [84.57, 86.18] | 89.68 [89.07, 90.28] | 59.09 [55.42, 62.71] | **89.39 [89.10, 89.68]** | **79.33 [78.39, 80.25]** | 83.52 | **69.21** | 78.75 |
| | Designed | 836 | **79.19 [76.49, 81.71]** | **88.91 [88.17, 89.68]** | **92.05 [91.60, 92.49]** | 61.12 [57.90, 64.33] | 88.64 [88.24, 89.03] | 74.69 [73.38, 75.99] | **87.20** | 67.91 | **80.77** |

Table S4. Results of our method compared with state-of-the-art baselines on the external verification task (Source: RIM-ONE-r3). All figures are DSC (%) with 95% CI. Average OD, OC and 'All' denote the simple average of OD, OC and all DSC.

| *Methods* | *Aug* | *Best step* | *Datasets* | | | | | | *Average* | | |
|---|---|---|---|---|---|---|---|---|---|---|---|
| | | | Target (Source: REFUGE) | | | | | | | | |
| | | | GoDARTS | IDRiD | Drishti-GS | | RIM-ONE-r3 | | | | |
| | | | OD | OD | OD | OC | OD | OC | OD | OC | All |
| nnU-Net | - | - | 66.21 [52.91, 78.34] | 43.61 [31.03, 56.31] | 95.88 [95.12, 96.54] | **83.40 [80.86, 85.73]** | 87.50 [86.09, 88.60] | 79.40 [75.34, 82.85] | 73.30 | **81.40** | 76.00 |
| DUNet | None | 54,948 | 50.84 [39.39, 61.82] | 71.79 [64.33, 78.49] | 78.92 [72.93, 84.01] | 56.53 [49.57, 62.81] | 53.24 [44.96, 61.17] | 40.24 [32.02, 48.46] | 63.70 | 48.39 | 58.59 |
| | Spatial | 4,476 | 52.94 [39.20, 66.41] | 78.37 [71.43, 84.60] | 86.88 [82.02, 90.83] | 58.56 [52.12, 64.54] | 66.55 [60.12, 72.34] | 50.39 [41.88, 58.60] | 71.19 | 54.48 | 65.62 |
| | Designed | 1,452 | 60.79 [49.07, 72.33] | 79.91 [74.69, 84.61] | 92.84 [89.82, 94.89] | 73.49 [69.55, 77.02] | 72.67 [66.46, 78.12] | 66.31 [59.47, 72.46] | 76.55 | 69.90 | 74.34 |
| TransUNet | None | 672 | 61.36 [51.61, 70.91] | 89.33 [85.50, 92.72] | 91.64 [89.73, 93.35] | 53.45 [49.66, 57.15] | 85.93 [84.32, 87.26] | 74.14 [70.07, 77.69] | 82.07 | 63.80 | 75.98 |
| | Spatial | 2,772 | 73.56 [66.85, 79.67] | 94.66 [93.42, 95.70] | 93.41 [91.32, 95.17] | 62.45 [59.17, 65.75] | 87.58 [86.69, 88.46] | 80.87 [76.52, 84.50] | 87.30 | 71.66 | 82.09 |
| | Designed | 888 | 71.05 [65.75, 76.52] | 90.83 [86.75, 94.11] | 94.13 [92.48, 95.51] | 73.72 [70.51, 76.73] | **88.44 [87.66, 89.21]** | 78.10 [74.93, 81.02] | 86.11 | 75.91 | 82.71 |
| Ours | None | 1,668 | 73.93 [67.47, 80.22] | 93.57 [92.19, 94.79] | 96.25 [95.68, 96.75] | 67.29 [63.00, 71.14] | 85.95 [84.60, 87.15] | 79.09 [75.81, 82.15] | 87.43 | 73.19 | 82.68 |
| | Spatial | 396 | **78.98 [73.54, 84.20]** | 93.71 [92.11, 95.05] | 96.06 [95.69, 96.43] | 76.34 [73.45, 79.04] | 84.81 [83.67, 85.89] | **81.81 [78.31, 84.94]** | 88.39 | 79.08 | 85.29 |
| | Designed | 3,684 | 77.41 [72.00, 82.48] | **94.91 [93.98, 95.78]** | **96.67 [96.29, 97.02]** | 76.29 [73.70, 78.75] | 87.70 [86.81, 88.55] | 80.79 [77.57, 83.74] | **89.17** | 78.54 | **85.63** |

Table S5. Results of our method compared with state-of-the-art baselines on the external verification task (Source: REFUGE). All figures are DSC (%) with 95% CI. Average OD, OC and 'All' denote the simple average of OD, OC and all DSC.

| *No.* | *Model* | *Size* | *Best step* | *Datasets* | | | | | *Average* |
|---|---|---|---|---|---|---|---|---|---|
| | | | | Source | Target | | | | |
| | | | | IDRiD | GoDARTS | Drishti-GS | RIM-ONE-r3 | REFUGE | |
| | | | | OD | OD | OD | OD | OD | All |
| 17 | RET + Seg(2) | 224 | 147,237 | 96.32 | 90.08 | 96.19 | 85.15 | 79.24 | 89.40 |
| 18 | Pre-A + Best + Post-A | 256 | 53,788 | 95.86 | 89.14 | 95.94 | 88.24 | 79.41 | 89.72 |
| 19 | Best + Skip | 256 | 2,600 | 96.87 | 90.98 | 96.51 | 87.57 | 91.02 | 92.59 |
| 20 | **FunduSegmenter** | 256 | 1,845 | 96.88 | 90.85 | 97.05 | 86.91 | 91.90 | **92.72** |

Table S6. Results of ablation study Group 2 experiments. All figures are DSC (%). Average 'All' denotes the simple average of all DSC. Acronyms used in this table denote the same meaning in Table 6.

| *Methods* | *Pre-processing* | *Best step* | *Datasets* | | | | | |
|---|---|---|---|---|---|---|---|---|
| | | | Source | Target | | | | |
| | | | GoDARTS | IDRiD | Drishti-GS | RIM-ONE-r3 | REFUGE | *Average* |
| | | | OD | OD | OD | OD | OD | All |
| DUNet | OD center crop | 3,595 | 75.35 | 94.12 | 91.72 | 73.84 | 90.79 | 87.62 |
| | Original | 21,520 | 95.88 | 97.69 | 89.82 | 18.58 | 86.91 | 73.25 |
| TransUNet | OD center crop | 920 | 65.05 | 88.59 | 82.17 | 89.22 | 91.21 | 87.80 |
| | Original | 15,800 | 95.82 | 97.63 | 90.64 | 62.35 | 90.71 | 85.33 |
| Ours | OD center crop | 205 | 78.99 | 95.41 | 95.54 | 88.73 | 91.31 | 92.75 |
| | Original | 910 | 95.20 | 96.66 | 92.88 | 91.56 | 91.70 | 93.20 |

Table S7. Results of ablation study Group 3 experiments. All figures are DSC (%). Average 'All' denotes the simple average of external experiment DSC.